\documentclass[10pt,twocolumn,letterpaper]{article}

\usepackage{iccv}
\usepackage{times}
\usepackage{epsfig}
\usepackage{graphicx}
\usepackage{amsmath}
\usepackage{amssymb}

\usepackage{multirow}
\usepackage{blindtext}
\usepackage{placeins}
\usepackage{adjustbox}
\usepackage{xfrac}
\usepackage{enumitem}
\usepackage{subcaption}

\usepackage{array}
\usepackage{rotating}
\usepackage{multicol}
\usepackage{chngcntr}
\usepackage{bm}
\usepackage{booktabs}

\usepackage{comment}
\usepackage{lipsum}
\usepackage{ifthen}

\usepackage[dvipsnames]{xcolor}

\usepackage[pagebackref=true,breaklinks=true,letterpaper=true,colorlinks,bookmarks=false]{hyperref}

\iccvfinalcopy  

\def\iccvPaperID{***} 

\begin{document}

\title{2nd Place Solution for SODA10M Challenge 2021 -
Continual Detection Track}

\author{Manoj~Acharya$^1$ \qquad Christopher Kanan$^{1,2,3}$\\
Rochester Institute of Technology$^1$, Paige$^2$, Cornell Tech$^3$\\
{\tt\small \{ma7583, kanan\}@rit.edu}\\
}

\maketitle
\ificcvfinal\thispagestyle{empty}\fi

\begin{abstract}
In this technical report, we present our approaches for the continual object detection track of the SODA10M challenge. We adapt ResNet50-FPN as the baseline and try several improvements for the final submission model. We find that task-specific replay scheme, learning rate scheduling, model calibration, and using original image scale helps to improve performance for both large and small objects in images. Our team `hypertune28' secured the second position among 52 participants in the challenge. This work will be presented at the ICCV 2021 Workshop on Self-supervised Learning for Next-Generation Industry-level Autonomous Driving (SSLAD).
\end{abstract}

\section{Dataset}
SODA10M~\cite{han2021soda10m} is a large-scale 2D object detection dataset specifically designed for autonomous driving. It consists of ten million unlabeled images with 20,000 fully annotated images for six categories: pedestrian, cyclist, car, truck, tram, and, tricycle. The labeled set is further split into 5K for training, 5K for validation, and the remaining 10K images for testing. These images are collected from 32 different cities under varying weather conditions and time to improve diversity.

The continual detection track comprises of four tasks with varying image domains as follows:\\
\textbf{Task 1:} Daytime, city-street and clear weather \\
\textbf{Task 2:} Daytime, highway and clear/overcast weather \\ 
\textbf{Task 3:} Night \\
\textbf{Task 4:} Daytime, rain

\begin{table}[b]
\centering
\caption{Training datasets for the four continual tasks along with instance count for six categories.}
\label{tab:trainset}
\resizebox{\linewidth}{!}{
\begin{tabular}{lccccccc}
\toprule
Task & Images & Pedestrain & Cyclist & Car   & Truck & Tram & Tricycle \\
\midrule
1 & 4470          & 4397       & 5885    & 21156 & 3870  & 1528      & 202      \\
2 & 1329          & 30         & 11      & 5161  & 4689  & 281       & 1        \\
3 & 1479          & 628        & 508     & 6224  & 1493  & 356       & 4        \\
4 & 524           & 84         & 285     & 1892  & 1234  & 154       & 5        \\
\bottomrule
\end{tabular}
}
\end{table}

\section{Introduction}
\label{sec:intro}

Object detection is concerned with localizing and classifying multiple categories in a scene. Current object detection methods are trained on large batches of fixed datasets without the expectation of any change during deployment. However, these assumptions do not hold anymore in real-world scenarios. For example, object detection in self-driving cars is much more difficult because the dataset domain can change e.g daytime to nighttime, clear to rainy, summer to snowy, etc. To deal with such challenges, static systems trained in an offline manner are not sufficient. Current detection models need to anticipate domain change and have the ability to continually adapt to new domains over time.

Another caveat of current object detection methods is that they are largely supervised. Human annotation of every objects to train the feature representation. However, this becomes inefficient as the number of classes is large, objects are small, and have objects with significant variations. One solution is to learn the representation in a self-supervised fashion. Self Supervised Learning learning (SSL) does not require any labels and the models are simply provided with large amounts of unlabelled data. Recent SSL approaches are competitive and perform close to their supervised counterparts with the only caveat being longer training times compared to the supervised models.

\section{Our Solutions}
\label{sec:model}
We use ResNet50-FPN~\cite{lin2017feature} model pre-trained on COCO  provided by the \textit{torcvision} library \cite{marcel2010torchvision} as the baseline model.
The Feature Pyramid Network (FPN) model uses features of multiple resolution and is better equipped to detect smaller object instances that are otherwise missed by backbones using only a single feature map. 

We use the Avalanche framework~\cite{lomonaco2021avalanche} for training and evaluating continual models. We train the models on an NVIDIA RTX A5000 (24 GB) GPU. The training time for a single model using a batch-size of eight images is roughly 12 hours.
We also use \textit{weights-and-biases} toolbox to monitor the loss curves and performance metrics during training. 

\begin{table*}[t]
\caption{ Model mAP scores for four tasks evaluated on the validation set. }
\centering
\label{tab:main-results}
\begin{tabular}{lllllc}
\toprule
Model                             & Task 1 & Task 2 & Task 3 & Task 4  &  Average \\
\midrule
Vanilla FPN - Offline             & 68.40   & 54.30   & 72.40   & 67.30   & 65.60      \\
Vanilla FPN - Fintetune           & 51.20   & 45.90   & 62.70   & 61.80   & 55.40      \\
\midrule
FPN, replay, \textit{lr} = 0.001, weight-decay =  $5e^{-5}$      & 53.70   & 44.00   & 65.80   & 60.70   & 56.05      \\
+large scale & 57.40   & \textbf{54.50 }  & 65.30   & 62.50   & 59.93      \\
+\textit{lr} = 0.02  & 58.70   & 45.20   & 69.20   & 74.60   & 61.93      \\
\midrule
FPN, replay, large scale, \textit{lr} = 0.01, no decay              & 59.30   & 47.00   & \textbf{69.50}   & 73.60   & 62.35      \\
+Temperature Scaling   &  \textbf{60.70}           &	48.30 &	69.30 &	\textbf{74.94} & \textbf{63.30} \\
\bottomrule
\end{tabular}
\end{table*}

\paragraph{Distillation:}
We use  distillation loss to minimize the difference between predictions from the original and the updated model. We select $N = 128$ boxes with the least background scores since they have a higher chance of having object instances. We sample 64 boxes to distill both bounding box regression outputs and the classification logits excluding the background class. Following  \cite{iccvkhmelkov}, we optimize the $L_2$ loss between ground truth and predictions along with the Faster RCNN loss to train the model. However, we find replaying raw examples performs better than distillation on the task sequences.

\paragraph{Replay:}
Instances of classes 1, 2, and 6 get rarer in the training sets after the first task. To ensure object instances for all classes are represented enough during training, we select images having the largest number of these rarer categories. Those images are then added to the memory buffer which are replayed during training. Since, the size of replay memory is limited to 250 images, we only add 83 images per task for the last three tasks.

\paragraph{Larger Image Scale:}
Images in the SODA dataset are of size 1920 $\times$ 1280 pixels. Torchvision library uses the default minimum and maximum sizes of 800 and 1333 pixels for object detection. We notice that smaller feature-map causes the model to miss or sometimes incorrectly classify bounding boxes. Training the model using the full image resolution improves the  average AP by roughly 4\% on the validation set.

\paragraph{Training details:}
We train the offline model for 10 epochs and other continual models for 10 epochs for each task. We use a Step Learning rate scheduler and reduced the learning rate by a factor of 10 after 6 epochs. We trained initial model versions with a learning rate of 0.001 and weight decay of 0.00005. After extensive tuning, we find that a learning rate of 0.01 with no weight decay achieves the best   validation performance of 62.35 average AP score.

\paragraph{Model Calibration:}
Visualizing the model predictions shows that our FPN detection model is overconfident and predicts softmax scores near 1.0 for many object instances. This produces lower AP values for higher precision intervals. Therefore, we soften the score distribution using a temperature scaling method and find that $T=2$ yields the best results.

\begin{table}[h]
\caption{Detection error categories with average reduction in AP(dAP) score generated by the TIDE \cite{bolya2020tide} toolbox.}
\label{table:tide}
\centering
\resizebox{\linewidth}{!}{
\begin{tabular}{lllllll}
\toprule
\textbf{Main Errors}   & \multicolumn{6}{l}{}                                               \\
Type                   & Classification  & Localization    & Both & Duplicate & Bkg  & Miss \\
dAP                    & 14.39           & 3.27            & 0.83 & 0.10      & 2.36 & 4.88 \\
\midrule
\textbf{Special Errors} & \multicolumn{6}{l}{}                                               \\
Type                   & False Positives & False Negatives &      &           &      &      \\
dAP                    & 9.71            & 23.14           &      &           &      &     \\
\bottomrule
\end{tabular}
}
\end{table}

\begin{table}[h]
\caption{Classification error matrix for the six object categories.}
\centering
\label{table:errormatrix}
\resizebox{\linewidth}{!}{
\begin{tabular}{lcccccc}
\toprule
\textbf{True class $\downarrow$ }          & Pedestrain & Cyclist & Car & Truck & Tram & Tricycle \\
\midrule
Pedestrain & 0          & 208     & 5   & 0     & 0          & 0        \\
Cyclist    & 322        & 0       & 40  & 13    & 2          & 12       \\
Car        & 7          & 53      & 0   & 408   & 99         & 1        \\
Truck      & 2          & 6       & 161 & 0     & 139        & 2        \\
Tram & 1          & 1       & 42  & 123   & 0          & 0        \\
Tricycle   & 8          & 19      & 9   & 6     & 0          & 0     \\
\bottomrule
\end{tabular}
}
\end{table}

\section{Conclusion and Recommendations}

Our best model obtains 63.30 average AP on the validation set and 55.67 average AP on the test set. We did not use the unlabelled set provided by the SODA10M dataset to pre-train our models. We believe using self-supervised pretraining on a large dataset can further improve the detection performance.

\begin{figure*}[!t]
\begin{center}
\includegraphics[width=1.0\linewidth]{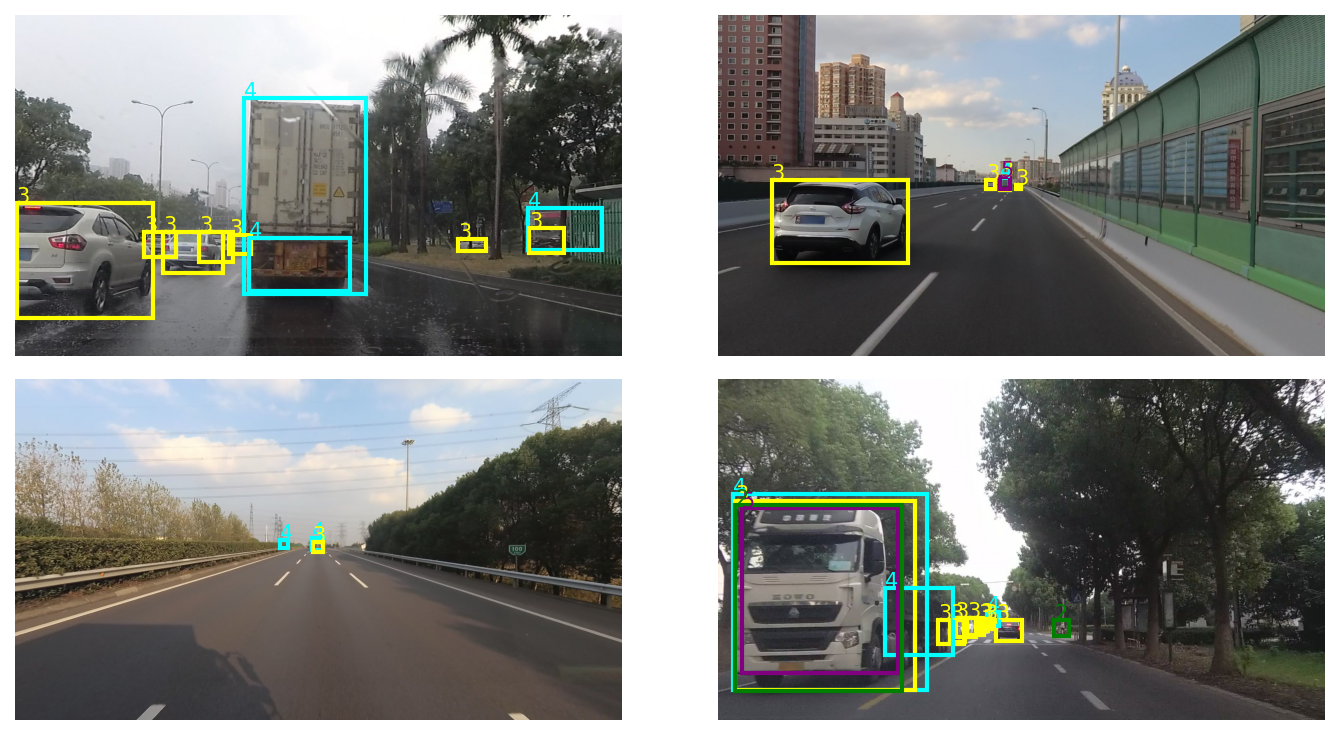}
\end{center}
  \caption{Predictions from our best model on images from the test split. Objects of category Truck(4) in cyan boxes are frequently confused with orange boxes representing category Car(3).}
\label{fig:test}
\end{figure*}

We analyzed the results using TIDE \cite{bolya2020tide} to analyze where the system is making errors and determine areas for further improvements. TIDE categorizes detector errors  into classification error, box localization error, missed detections, duplicate predictions and background confusion as shown in Table \ref{table:tide}. The largest error source is classification error which occurs when the model predicts box with incorrect class but with high IOU score with the ground truth box. Fixing classification errors alone can improve the overall mAP by 14.39 \%. The error matrix in Table \ref{table:errormatrix} shows that the most confused classes are pedestrians which are confused for cyclists and cars confused for trucks. It is not unusual given the relative visual similarity between the categories plus differentiation is difficult for bounding boxes of low resolution. Further improvements can be gained by fixing all other types of errors.

Another area of improvement is to use better backbone models. Also, newer detection architectures such as Double-Head \cite{wu2020rethinking} have been shown to improve localization performance that can improve the overall mAP score. Similarly, Cascade-RCNN \cite{cai2018cascade} filters bad proposals though multiple stages to  produce higher quality bounding boxes.

Data augmentation can play a huge role in the overall performance gain. We only use random horizontal flips augmentation for all our models. Smart augmentation strategies such as thumbnail augmentation \cite{cutpaste} can be used to increase performance for rarer categories. 

In the future, models that can adapt to any domains without explicit training are necessary. Recent works in universal and continual domain adaptation \cite{wulfmeier2018incremental,you2019universal} have shown promising results on image classification tasks. Finally, in this work, we simply replay images having the largest number of rarer classes. Better replay strategies \cite{acharya2020rodeo} can further help to improve the efficiency and the performance of continual detection systems.

{\small
\bibliographystyle{ieee_fullname}
\bibliography{egbib}
}

\end{document}